# Eradicating Social Biases in Sentiment Analysis using Semantic Blinding and Semantic Propagation Graph Neural Networks

**Hubert Plisiecki**

hplisiecki@gmail.com

Institute of Psychology, Polish Academy of Sciences

Society for Open Science

ORCID: 0000-0002-5273-1716

## Abstract

This paper introduces the Semantic Propagation Graph Neural Network (SProp GNN), a machine learning sentiment analysis (SA) architecture that relies exclusively on syntactic structures and word-level emotional cues to predict emotions in text. By semantically blinding the model to information about specific words, it is robust to social biases such as political or gender bias that have been plaguing previous machine learning-based SA systems. The SProp GNN shows performance superior to lexicon-based alternatives such as VADER (Valence Aware Dictionary and Sentiment Reasoner) and EmoAtlas on two different prediction tasks, and across two languages. Additionally, it approaches the accuracy of transformer-based models while significantly reducing bias in emotion prediction tasks. By offering improved explainability and reducing bias, the SProp GNN bridges the methodological gap between interpretable lexicon approaches and powerful, yet often opaque, deep learning models, offering a robust tool for fair and effective emotion analysis in understanding human behavior through text.

**Eradicating Social Biases in Sentiment Analysis using Semantic Blinding and Semantic Propagation Graph Neural Networks**

The automated assessment of emotional content in textual data, or Sentiment Analysis (SA) has revolutionized research across the social sciences, enabling applications such as suicide risk prediction from text messages (Glenn et al., 2020), analysis of historical well-being trends (Hills et al., 2019), political election forecasting (Ramteke et al., 2016), and monitoring global emotional responses during crises like the COVID-19 pandemic (Wang et al., 2022). Thanks to SA researchers gained access to an extensive array of authentic data on human emotions, as vast as the multitude of texts available on the internet.

However, current methods for emotion assessment have notable limitations. Transformer-based architectures, and other machine learning models — while being recommended for their high performance (Widmann & Wich, 2022), are at the same time susceptible to inheriting social biases from their training data, including gender, racial, ageist, and political biases (Kiritchenko & Mohammad, 2018; Díaz et al., 2018; Plisiecki et al., 2024). For instance, Kiritchenko & Mohammad surveyed 219 automatic sentiment analysis systems 75% of which showed signs of significant racial and/or gender bias. A more targeted investigation conducted found that a model trained on annotated political texts exhibited biases aligned with the political orientation of the annotators. Removing bias relevant items from the training data reduced these biases, implying that the annotations were their source of origin (Plisiecki et al., 2024). As these findings highlight, addressing bias in emotion modeling has become an essential challenge for sentiment analysis research.

Given that balancing the annotator group in terms of bias is problematic as, aside from the labor required to find the right people, there always exists a risk of the existence of a bias that was not accounted for. The alternative so far has been the use of simpler models, such as those based on lexicons (also called norms or dictionaries), which are long lists of manually selected words annotated for their emotional information (Plisiecki et al., 2024). The most basic lexicon approaches rely on simply looking up the emotional value of each available word in a text and averaging the results. Unfortunately, this approach works well only for very simple texts, as it does not consider syntactic information. An example here is negation, which can transform the meaning of a word in a sentence but goes unnoticed by simple dictionaries.

More complex alternatives rely on hard coded rules to handle syntactic dependencies. Examples of such approaches are the VADER (Valence Aware Dictionary and Sentiment Reasoner), and EmoAtlas (Hutto & Gilbert, 2014; Semeraro et al., 2023). Both of these techniques rely on dictionaries combined with hard-coded rules that were arrived at through examination of sentence structures. Rules such as "if the negation is three words away from an emotionally loaded term, flip the emotional loading of the term" allow those models to handle negations and other semantic structures beyond the reach of normal lexicons. Their performance however rarely approaches that of pretrained transformers and the degree of generalization to languages different than English is questionable, due to different syntactic patterns being present in languages further away on the language phylogenetic tree.

**The Proposed Solution**

To provide a better solution, this paper presents the Semantic Propagation Graph Neural Network (SProp GNN), a supervised approach that bridges the methodological gap between simple lexicon-based methods and complex black-box models providing high performance that is robust to training data bias. This approach uses the syntactic relationships within sentences to create graphs enhanced with emotion information at the word level. The SProp GNN is then trained on these graphs, providing emotion predictions at text level. The risk of bias propagation is reduced by purposefully blinding the model to semantic information that it could otherwise overfit.

The SProp GNN emotion prediction pipeline can be split into three distinct stages (as seen on Figure 1.):

**Stage A – Word Level Emotion Prediction** - The emotional value of each word is identified.

**Stage B – Syntactic Graph Creation** – The sentence is transformed into a graph that reflects the syntactic connections between words.

**Stage C – Semantic Propagation Graph Neural Network** – A specialized neural network processes these graphs, along with the emotional information of singular words, to predict the overall emotional meaning of a text, without relying on the direct knowledge of the words that constitute it.

Figure 1. *Steps of the Emotion Prediction Pipeline*

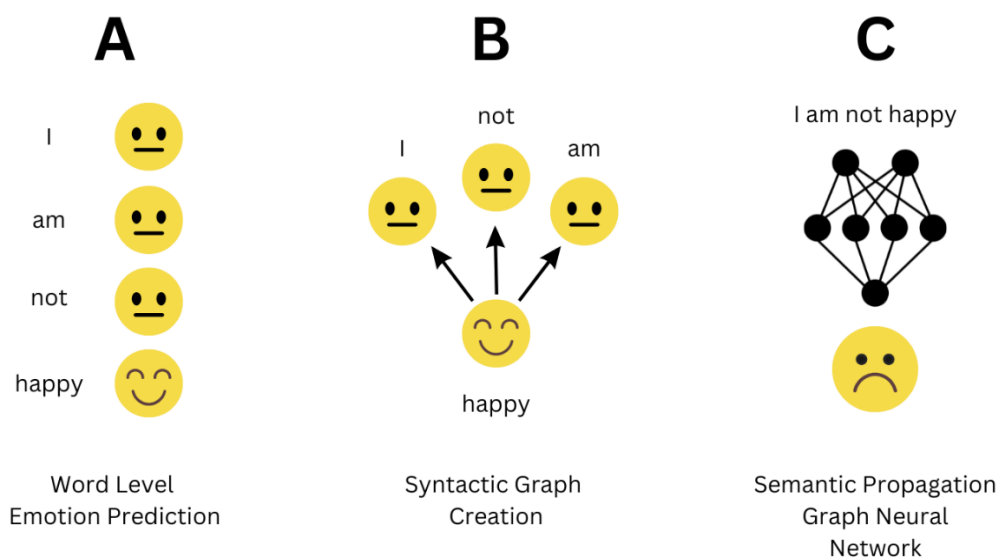

The proposed approach of selectively withholding specific semantic information from the model can be termed *Semantic Blinding*. Deliberately limiting the model's access to particular semantic details prevents it from associating emotional predictions with specific words or concepts that could introduce unwanted biases. By focusing exclusively on syntactic relationships and word-level emotional cues, semantic blinding ensures that the model's emotional assessments are free from training data biases related to specific groups or subjects. This technique, therefore, enhances the model's capacity to generalize across varied text sources without inheriting unintended, potentially harmful associations, providing a robust and unbiased tool for emotion prediction.

**Brief Introduction to Graph Neural Networks**

Graphs are mathematical structures used to represent entities (nodes) and their relationships (edges), providing a powerful framework for analyzing structured data (Zhou et al., 2020). In the context of natural language processing (NLP), a sentence can be represented as a graph where words serve as nodes, and edges capture syntactic or semantic relationships, such as dependencies between words. Graph Neural Networks (GNNs) extend this framework by applying deep learning techniques to graphs, enabling models to learn from the relationships and structures inherent in the data. Unlike traditional neural networks, which operate on fixed-size inputs like vectors or grids, GNNs analyze the connectivity patterns and features of nodes and edges to perform tasks such as classification, prediction, or clustering (Zhou et al., 2020). For example, GNNs have been successfully used in applications ranging from molecule property prediction in chemistry to fraud detection in financial networks (Motie & Raahemi, 2024; Wieder et al., 2020). They have also gained recognition in sentiment analysis applications (for a comprehensive review see Rad et al., 2023), however since the main aim of previous methods was to maximize the predictive performance of their approaches, none of them limited the amount of information that the model received, as is done in the case of Semantic Blinding.

**The Contents of the Paper**

Through comparative experiments, this paper demonstrates that the SProp GNN outperforms traditional lexicon-based models as well as lexicon-based alternatives across both discrete and dimensional emotion prediction tasks in English and Polish. It closely approaches transformer-based model accuracy in both languages, and task types offering a compelling alternative to biased black-box models. Furthermore, the paper provides detailed statistical and theoretical evidence that the SProp GNN is robust to the biases shown in previous research.

<div align="center">**Methods**</div>

**The Emotion Prediction System**

The method proposed by the current paper essentially combines the use of three different machine learning based approaches, which process the text sequentially.

*Word Level Emotion Prediction*

The task of the first model is to predict the emotional value of the words in the text. This task, also seen as norm-, or lexicon-extrapolation is currently best attempted using transformer-based models (Plisiecki & Sobieszek, 2023). For the purposes of the current paper, either existing pretrained transformers norm extrapolation models are used, or new ones are trained when no off-the-shelf solutions are available. This stage results in emotion estimates for each separate word in a given text.

*Creation of the Syntactic Graph*

The text is then divided into sentences, and these sentences are analyzed using the *spaCy* package (Ines Montani et al., 2023). This software uses machine learning algorithms and linguistic rules to parse text, creating detailed syntactic structures for each sentence. *SpaCy* generates dependency graphs, which represent the relationships between words, as well as dependency labels (e.g. negations) and part-of-speech (POS) tags (e.g. verb). If a text consists of multiple sentences, these are linked back together using dedicated sentence nodes. This procedure allows the framework to capture the structure of each text, providing a type of scaffolding for the SProp GNN to propagate word-level emotional information through. This stage results in the creation of a syntactic graph, enriched with the information about the part-of-speech categories, and emotions from the first stage, at the node level, and dependency labels at the edge level.

*Semantic Propagation Graph Neural Network*

The final part of the pipeline is the SProp GNN, a neural network model designed to propagate emotional information extracted in the first stage of the pipeline, through the graph generated in the second stage. SProp GNN can be split into three main components: a custom SProp (GNN) layer, an attention pooling mechanism, and linear output layers.

**Custom Graph Neural Network Layer**: At the heart of the model is the custom Semantic Propagation Layer. This layer operates on the dependency graphs generated by spaCy (Ines Montani et al., 2023), where each node represents a word with associated features, and edges represent syntactic relationships between words. The Semantic Propagation Layer integrates information from the word node features (earlier predicted emotional load, and part-of-speech tags) and edge features (such as dependency types) to compute a scaling factor for each of its edges. It then propagates the emotional information from word nodes along those edges, scaling them accordingly. The hope here is that by doing so, it can model the propagation of emotional information through the sentence.

**Attention Pooling Mechanism**: After the graph has been processed by the SProp GNN layer, the model employs an attention-based pooling mechanism. This component aggregates the information from all nodes in the graph to create a single, fixed-size vector representation of the entire text. The attention mechanism assigns different weights to word nodes based on their relevance, effectively allowing the model to focus on the most significant words and relationships when forming this overall representation.

**Linear Output Layers**: The aggregated text representation is then passed through multiple linear layers. These layers transform the high-dimensional embedding into a scalar value between 0 and 1, corresponding to the predicted score for each predicted emotional metric. By having separate output layers for each metric, the model can simultaneously make multiple predictions, each tailored to the unique aspects of the respective psychological construct. This is different for discrete classification, where the layers transform the embedding to a vector of size equal to the number of predicted classes. This vector, when transformed, becomes an array of class probabilities.

The SProp GNN model processes text by first constructing a rich representation of its syntactic and semantic structure using the SProp layer. It then distills this information into a concise and meaningful summary via attention pooling. Finally, it translates this summary into actionable predictions through the linear output layers. Before prediction, this model has to be trained on a dataset of texts, with annotated emotional metrics in the form of either emotion intensities, or discrete emotion classes. As the model does not have direct access to the words with regards to which people exhibit social bias (e.g. certain politicians, gender information etc.), it cannot learn the association between them and the biased emotion estimates. Therefore, the biased part of an emotion estimate is from its perspective indistinguishable from noise, as it has no systematic relationship with the input data. This renders the model blind to the socially sensitive features of the input, therefore rendering it agnostic with regards to social biases. For a more detailed description of the model architecture see the Technical Appendix.

## Comparative Experiments

The SProp model has been tested on three separate datasets, the GoEmotions dataset, the EmoBank dataset, and the dataset used in the Plisiecki and colleagues political bias study, referred to from now on as the Polish Political Dataset (2024). These datasets cover two languages (Polish and English), and two different emotion prediction tasks (categorical, and continuous emotion prediction).

### *The GoEmotions Dataset*

The GoEmotions dataset, developed by Google researchers (Demszky et al., 2020), consists of around 58,000 English Reddit comments annotated with 28 distinct emotions, totaling over 210,000 annotations. Sourced from a Reddit data dump spanning 2005 to early 2019, the dataset includes comments from diverse subreddits, balanced by capping comment counts from the most popular communities and sampling evenly across others. Emotion categories were curated based on psychological research to represent a broad but non-overlapping range of emotions. Each comment received annotations from three English-speaking raters from India, with additional raters assigned when agreement was low.

### The EmoBank Dataset

The EmoBank dataset, created by Buechel and Hahn (2022), consists of 10,062 English sentences from sources like news, blogs, fiction, and letters, annotated along three emotional dimensions: Valence, Arousal, and Dominance (VAD). Each sentence was rated by five annotators from the crowdsourcing platform CrowdFlower for both *writer* and *reader* perspectives on a 5-point scale, giving insights into both expressed and perceived emotions. In accordance with the recommendations of the researchers, the current paper uses the version of the dataset with the weighted average of the reader and writer perspective labels provided at their online repository (*JULIELab/EmoBank*, 2017/2024).

### The Polish Political Dataset

The Polish Political dataset (Plisiecki et al., 2024) includes 1.25 million Polish social media posts from journalists, politicians, NGOs, and general users. The emotionally neutral texts were filtered out using lexical norms on valence, arousal, and dominance. The final 10,000 texts were annotated by 20 psychology-trained annotators on six emotions (happiness, sadness, anger, disgust, fear, and pride) and two dimensions (valence and arousal) using a 5-point scale. Each annotator completed five weekly sets of 100 randomly assigned texts, ensuring each text was labeled by five raters for reliable coverage and minimizing cognitive fatigue over the five-week process. The resulting scores were averaged to create an intensity score for each text – emotion pair.

### Dataset Preparation

Each of the dataset was first prepared by either calculating the most voted emotion category in the case of GoEmotions or normalizing the intensity of annotations to 0 to 1 range in the case of the two continuous datasets. Each dataset was then split into the training, evaluation, and test subsets in a proportion of 8:1:1, with the exception of the Polish dataset, for which the split dataset was taken from the original paper (Plisiecki et al., 2024). For more information about the preparation of the datasets and the datasets themselves see the Technical Appendix.

## Comparative Approaches

The aforementioned datasets are used to compare the SProp model's performance to four alternative methods. The first three methods rely on lexicons, and as such are resilient to annotator bias. In order for the proposed framework to become a preferred alternative to them, it has to outperform them on evaluation metrics. The fourth method relies on transformer base models to predict emotions. It is added for comparison with high performing, but bias prone, models to better inform researchers' decision-making. Each of the methods' performances is calculated on the test sets of respective datasets.

### The Lexicon Approach

The lexicon approach works by averaging the emotional intensity of words in a given text. In the case of the Emobank dataset and the Polish political dataset I average the word ratings of previously published transformer-based norm extrapolation models (Plisiecki et al., 2024). In the results section I only report the results of averaging after removing stop words, as this method attained better results. As the EmoAtlas approach has proven superior to the lexicon approaches (Semeraro et al., 2023) in the task of discrete emotion prediction on the GoEmotions dataset, I do not report the performance of the lexicon approach for that specific task.

### The Vader Approach

VADER (Valence Aware Dictionary and Sentiment Reasoner) is a rule-based model designed for sentiment analysis, particularly effective in capturing sentiment from social media and informal text. VADER combines a lexicon with rules that account for various intensifiers, negations, and punctuation, making it particularly adept at assessing the sentiment intensity conveyed in short online texts. VADER assigns polarity scores for positive, neutral, and negative sentiment, averaging these scores to produce an overall sentiment value for a given text. This approach is only capable of producing valence estimations (Hutto & Gilbert, 2014).

### The EmoAtlas Approach

The EmoAtlas utilizes an extensive lexicon-based network to profile emotions by mapping syntactic and semantic relationships in text, effectively capturing nuanced emotional cues without extensive model training. Using validated emotional lexicons for Plutchik's eight core emotions in conjunction with a spaCy based (Ines Montani et al., 2023) syntactic analysis, it efficiently identifies emotional tones in multiple languages. Its rule-based structure enables it to run significantly faster than transformer-based models, providing researchers with interpretable insights into how emotions are conveyed in text associations (Semeraro et al., 2023).

### The Transformer Approach

A base transformer can be finetuned to predict both continuous and categorical emotions. Here the *roberta-base* transformer model developed by Facebook (Liu et al., 2019) is finetuned on the two aforementioned English datasets. After a hyperparameter sweep for each dataset, the final models were trained on the parameter setup that led to the best performance. For the Polish political bias dataset, the performance of the GNN model is compared with a transformer model that was finetuned to predict emotions in the original dataset paper (Plisiecki et al., 2024). For a more detailed description of the implementation of each of the comparative approaches refer to the Technical Appendix.

### Testing for Bias

This section evaluates whether the SProp GNN model mitigates overfitting on biases present in the training data. It follows the approach established by Plisiecki et al. (2024) [1], who analyzed a transformer-based language model's predictions of sentiment toward 24 prominent Polish politicians. The original study selected politicians based on a public trust survey and tested the model's sentiment responses to three types of text stimuli: (1) the politicians' names alone, (2) politically charged sentences containing these names, and (3) neutral sentences featuring the same names. The model's task was to classify each stimulus as having positive or negative valence.

To quantify political bias, Plisiecki et al. (2024) fit linear regression models in which the model's predicted valence was the dependent variable (Y), and the politician's political affiliation (a dummy-coded factor) and gender were predictors (X1: political affiliation, X2: gender). Thus, the model took the form:

$$Y = \beta_0 + \beta_1(\text{political affiliation}) + \beta_2(\text{gender}) + \epsilon$$

They found that these predictors explained a substantial proportion of the variance in valence predictions (52% for neutral sentences, 66% for political sentences, and 67% for names alone), indicating significant bias.

The current study replicates and extends this approach with the SProp GNN model using three complementary methods. The goal is to determine whether SProp GNN exhibits significantly less bias than the previous model.

### Approach 1: Replicating the Original Regression Procedure

The first approach directly replicates the original regression methodology. The same linear regression model is applied:

$$Y_{\text{SProp}} = \beta_0 + \beta_1(\text{political affiliation}) + \beta_2(\text{gender}) + \epsilon$$

Here, $Y_{\text{SProp}}$ represents the valence predictions made by the SProp GNN for each stimulus. The null hypothesis (H0) states that the SProp GNN model does not exhibit bias (i.e., $\beta_1 = \beta_2 = 0$) while the alternative hypothesis (H1) is that at least one of the bias coefficients is non-zero.

To test this, a permutation test on the observed valence predictions is employed. The correspondence between stimuli and the predictor values (political affiliation, gender) is randomly shuffled 100,000 times and the regression is re-estimated each time. This produces a null distribution of test statistics (e.g., F-statistics or sums of squared residuals) (Manly, 1997). If the observed statistic falls into the extreme tails of this distribution, the H0 is rejected with the conclusion that the SProp GNN exhibits bias. Non-significant results should be interpreted with caution, as it is difficult to ascertain the test's power precisely.

[1] The author thanks Paweł Lenartowicz for help in coming up with the statistical tests required to test the SProp model's bias

### Approach 2: Assessing Bias Reduction in the SProp Model

The second approach aims to determine whether the SProp GNN model reduces bias compared to the original model. Instead of testing if bias exists, the test related to whether the SProp model's bias is equivalent to (or less than) that of the original transformer model.

First, the SProp GNN predictions are adjusted by removing the estimated bias from the original model. To do this, the original model's estimated bias coefficients are used ($\widehat{\beta_1}$ and $\widehat{\beta_2}$) to create adjusted predictions:

$$Y_{\text{SProp, adjusted}} = Y_{\text{SProp}} - \left(\widehat{\beta_1} \cdot \text{political affiliation} + \widehat{\beta_2} \cdot \text{gender}\right)$$

Next, a regression is performed with the original bias factors as predictors on these adjusted scores. If the adjusted SProp predictions still show a significant relationship with the bias factors, it means the SProp model retained the same pattern of bias. If the bias factors do not predict the adjusted valence (or predict it inversely), it suggests bias has been reduced.

The null hypothesis ($H_0$) for this approach states that the SProp model's bias is the same as the original model's bias. The alternative hypothesis ($H_1$) suggests that the SProp model's bias is reduced. This is tested using a one-sided permutation test (100,000 random assignments). A statistically significant negative beta coefficient would indicate that the SProp model is inversely related to the original bias factor, signifying bias reduction.

### Approach 3: Comparing Differences Between Models

The third approach examines the difference in predictions between the transformer model and the SProp model. Define the difference in predicted valence as:

$$\Delta Y = Y_{\text{SProp}} - Y_{\text{transformer}}$$

This difference is then regressed on the original bias factors:

$$\Delta Y = \gamma_0 + \gamma_1 (\text{political affiliation}) + \gamma_2 (\text{gender}) + \epsilon$$

The null hypothesis ($H_0$) is that the difference in predictions between models is unrelated to the bias factors ($\gamma_1 = \gamma_2 = 0$). The alternative hypothesis ($H_1$) is that these factors significantly predict the difference, confirming that bias is driving the disparities between models.

As before, a one-sided permutation test (100,000 random assignments) is conducted to determine whether the observed association differs from what would be expected by chance. A significant result would indicate that bias plays a key role in differentiating the two models.

### Note on Interpreting the Results

While these methods help determine whether the SProp GNN model reduces bias, it is important to recognize that errors in the original model's estimated bias parameters may attenuate the observed relationships in the SProp model. Due to such estimation errors, the bias parameters in the new tests are not expected to reach exactly 1 (or to show a perfect elimination of bias). Even if no bias were present in the SProp model, random measurement errors and attenuation effects may prevent the parameters from perfectly reflecting the removal of bias.

## Explainability

For the sake of explainability, the SProp GNN saves its scaling factors as well as attention weights, allowing the user to understand the type of information on which the model based its decisions. As a full analysis of how the model reacts to a large array of diverse sentences is beyond the scope of this paper, the focus is shifted to explaining the basic mechanics using two sentences, employing an emotional word and a negation: "I am happy" and "I am not happy." The activity of the model is then compared between these sentences.

This explainability approach is particularly important because it allows users to assess not only the model's outputs but also the reasoning behind them. By exposing the scaling factors and attention weights, it becomes possible to pinpoint how specific words and their relationships, such as the negation in "I am not happy," influence the emotional predictions. This transparency is crucial in ensuring trust and interpretability in sentiment analysis models, especially for applications where ethical considerations or fairness are paramount.

## Results

### The GoEmotions Dataset

In the task of discrete emotion prediction conducted on the GoEmotions dataset, the Semantic Propagation GNN generally outperforms the EmoAtlas approach across the three key performance metrics: accuracy, precision, and recall (See Table 1.). Here, accuracy measures how often the model's predictions are correct overall, precision assesses the proportion of correct positive predictions among all positive predictions made, and recall evaluates the model's ability to identify all actual instances of each emotion. The only exceptions are in the precision metric for the emotions of anger and disgust, where Emo Atlas slightly exceeds the GNN.

While the SProp GNN shows better performance than Emo Atlas, both methods are generally surpassed by their transformer-based counterpart. The RoBERTa model, which leverages advanced language representations, outperforms both the Emo Atlas and the Semantic Propagation GNN across all emotions and metrics. However, the GNN is not far behind RoBERTa, achieving a mean accuracy difference of only 5.70 percentage points, compared to a difference of 20.73 percentage points between RoBERTa and the Emo Atlas.

Similarly, for precision, the average difference between RoBERTa and the GNN is 17.05 percentage points, whereas the difference between RoBERTa and the Emo Atlas is 30.33 percentage points. In terms of recall, the average difference between RoBERTa and the GNN is 20.81 percentage points, while the difference between RoBERTa and the Emo Atlas is 48.46 percentage points. These results indicate that while the GNN approaches the performance of RoBERTa, the Emo Atlas method lags significantly behind in all three metrics.

**Table 1.**

*Performance results on the goemotion dataset*

| emotion | Accuracy score % | | | Precision score % | | | Recall score % | | |
|---|---|---|---|---|---|---|---|---|---|
| | roberta | emoa | sprop | roberta | emoa | sprop | roberta | emoa | sprop |
| anger | 91.3 | 70.0 | **80.0** | 86.7 | **70.1** | 65.9 | 87.8 | 33.8 | **82.3** |
| disgust | 93.9 | 66.8 | **88.9** | 78.2 | **73.7** | 50.7 | 63.0 | 19.9 | **35.1** |
| fear | 94.7 | 77.3 | **93.2** | 71.2 | 39.6 | **72.1** | 85.6 | 48.2 | **59.6** |
| joy | 97.3 | 73.7 | **92.2** | 93.0 | 70.2 | **76.4** | 90.7 | 47.5 | **77.8** |
| sadness | 94.8 | 71.6 | **88.9** | 81.4 | 52.3 | **61.5** | 80.2 | 35.5 | **48.9** |
| surprise | 96.1 | 84.3 | **90.7** | 85.1 | 7.7 | **66.7** | 86.4 | 18.0 | **65.1** |

*Note.* The emotion categories had to be limited to those presented in the table both due to lexicon availability and EmoAtlas emotion coverage. The results written in bold pinpoint the best performance in a given metric out of the two alternatives to transformers. The metric results for the EmoAtlas were taken from the original manuscript, which introduced the technique. While that means that they were tested on a wider test set, it still provides a good overview of the approach performance given that it does not require any finetuning. Model codes: roberta – finetuned transformer; emoatlas – the Emo Atlas approach; sprop – Semantic Propagation GNN model.

**The EmoBank Dataset**

The task of sentence level emotion prediction was run on the EmoBank dataset, on which the SProp GNN outperformed both the lexicon approach and the Vader approach (see Table 2.). While RoBERTa achieved higher scores than the SProp GNN, the degree of difference varied between predicted metrics. While in the case of valence the difference amounted to 0.13 points, for arousal it was as low as 0.02.

**Table 2.**

*Performance results on the emobank dataset*

| Metric | roberta | lexicon | vader | sprop |
|---|---|---|---|---|
| Valence | 0.75 | 0.45 | 0.46 | **0.62** |
| Arousal | 0.48 | 0.25 | - | **0.45** |

*Note.* The results written in bold pinpoint the best performance, measured using the Pearson's correlation, in a given metric out of the three alternatives to transformers. Model codes: roberta – finetuned transformer; vader – the Vader approach; lexicon – the lexicon approach; sprop – Semantic Propagation GNN model. The lexicon score has been calculated after pruning stopwords.

**The Polish Political Dataset**

Finally, the Polish political dataset was used to test the model performance on a mixed set of both multiple and single sentence texts. Here, the SProp outperformed its lexicon counterpart yet again (See Table 3.). Unsurprisingly it was at the same time worse at predicting emotion scores than the RoBERTa model by 0.16 points in the case of valence, and 0.13 points in the case of arousal.

**Table 3.**

*Performance results on the polish political dataset (Pearson's Correlation)*

|         | roberta | lexicon | sprop    |
|---------|---------|---------|----------|
| Valence | 0.88    | 0.57    | **0.72** |
| Arousal | 0.75    | 0.33    | **0.62** |

*Note.* The results written in bold pinpoint the best performance in a given metric, measured using the Pearson's correlation, out of the two alternatives to transformers. Model codes: roberta – finetuned transformer; lexicon – the lexicon approach; sprop – Semantic Propagation GNN model. The lexicon score has been calculated after pruning stopwords.

**Political Bias Results**

***Approach 1: Replicating the Original Regression Procedure***

The replication of regressions performed in the original bias study (Plisiecki et al., 2024) yielded no significant results for the SProp GNN model (see Table 4). Neither political affiliation nor gender explained any meaningful variance in the model's valence predictions across any of the stimuli categories: names, neutral sentences, and political sentences. This outcome contrasts sharply with the results obtained for the transformer model in the original study, where political affiliation and gender were significant predictors, explaining 66% of the variance in valence for political sentences, 52% for neutral sentences, and 67% for names alone. For the SProp model, these same predictors explained a negligible proportion of the variance, as shown by the $R^2$ values of 0.077, 0.135, and 0.103, which are accompanied by non-significant permutation test p-values.

In Table 4, the results are presented for both models. The regression intercepts represent the predicted valence for the reference group, which is Zjednoczona Prawica (the ruling party) and male politicians, while the coefficients for political affiliation indicate how much valence changes for other groups (e.g., Konfederacja, Koalicja Obywatelska, etc.). For the transformer model, the political affiliation coefficients are consistently significant across all stimuli types, confirming substantial bias in its predictions. For example, the valence associated with Koalicja Obywatelska is consistently higher than that for Zjednoczona Prawica, with coefficients such as 5.83 (neutral sentences) and 2.30 (names only), both significant at $p < 0.05$. Gender also has a notable influence in the transformer model, with a coefficient of 9.77 for neutral sentences, indicating that valence predictions for women are substantially more positive than for men in this category.

In contrast, the SProp GNN model shows no significant coefficients for political affiliation or gender in any stimulus category. For instance, the coefficient for Koalicja Obywatelska is close to zero (e.g., 0.18 for neutral sentences, 0.93 for names, and 0.60 for political sentences) and accompanied by p-values well above 0.05. Similarly, the gender coefficient is 3.19 for neutral sentences, 2.11 for political sentences, and while as high as 8.10 for names only, it is not statistically significant. These results suggest that the SProp GNN model's predictions are less systematically influenced by political affiliation and gender, highlighting a potential reduction in bias compared to the transformer model.

Despite these findings, it is important to interpret the results with caution. While the lack of statistical significance in the SProp GNN model suggests an absence of systematic bias, this alone does not confirm that the model is entirely unbiased. The low explanatory power of the regressions and the non-significant results may also reflect limitations in the sensitivity of the statistical tests or the sample size, rather than the true absence of bias. Moreover, differences in residual variance and standard errors between the models indicate that additional factors may be influencing the outcomes. Therefore, complementary analyses, such as those presented in later sections, are essential to provide a more comprehensive understanding of bias reduction in the SProp GNN model.

**Table 4.**
*Regression models – differences in valence*

|  | Transformer | | | Semantic Propagation GNN | | |
|---|---|---|---|---|---|---|
|  | *Dependent variable: Valence of:* | | | | | |
|  | Names only (1) | Neutral Sentences (2) | Political Sentences (3) | Names only (1) | Neutral Sentences (2) | Political Sentences (3) |
| intercept | 45.40*** | 48.61*** | 43.89*** | 53.59*** | 55.02*** | 44.96*** |
|  | (0.67) | (0.92) | (0.26) | (1.87) | (0.71) | (0.49) |
| 3D | 5.73** | 9.09** | 2.72** | 2.33 | 1.44 | 0.73 |
|  | (2.37) | (3.26) | (0.93) | (6.61) | (2.52) | (1.71) |
| K | 6.15* | 8.37* | 2.56* | 3.85 | 0.83 | 1.02 |
|  | (3.10) | (4.28) | (1.22) | (8.67) | (3.30) | (2.24) |
| KO | 5.83*** | 3.71* | 2.30** | 0.93 | 0.18 | 0.60 |
|  | (1.31) | (1.80) | (0.51) | (3.66) | (1.39) | (0.94) |
| Left | 3.03 | 5.46 | 2.51** | -3.75 | -3.02 | -1.14 |
|  | (2.56) | (3.53) | (1.01) | (7.17) | (2.73) | (1.85) |
| gender | 9.77** | 10.10* | 1.88 | 8.10 | 3.19 | 2.11 |
|  | (3.48) | (4.80) | (1.37) | (9.74) | (3.71) | (2.52) |
| Observations | 22 | 22 | 22 | 22 | 22 | 22 |
| $R^2$ | 0.665 | 0.520 | 0.662 | 0.077 | 0.135 | 0.103 |
| Adjusted $R^2$ | 0.547 | 0.370 | 0.556 | -0.211 | -0.135 | -0.178 |
| Residual Std. Error | 3.38 (df=16) | 5.21 (df=16) | 1.29 (df=16) | 6.69 (df=16) | 2.75 (df=16) | 1.70 (df=16) |
| P-value (permutation) | 0.008*** | 0.049** | 0.018** | 0.885 | 0.686 | 0.793 |

*Note.* *p<0.1; **p<0.05; ***p<0.01 (*t*-test)
Zjednoczona Prawica (ruling party) as intercept, gender: woman=1, man=0
*Abbreviations*: K – Konfederacja, 3D – Trzecia Droga, KO – Koalicja Obywatelska, Left – Nowa Lewica

### Approach 2: Assessing Bias Reduction in the SProp Model

The beta coefficient for bias was $\beta$ = -0.78, with a permutation p-value significant at $p$ = 0.012. This result suggests that the SProp model's valence predictions exhibit a substantially lower level of bias compared to those of the transformer model. The negative value of the coefficient indicates an inverse relationship, suggesting that the bias introduced by political affiliation and gender in the original model has been largely mitigated in the SProp GNN model. Given the potential for real-world measurement variability and the effects of regression dilution, a coefficient of -0.78 strongly implies that the SProp GNN has no significant residual bias from the original model, or that any remaining bias is minor and unlikely to have practical significance.

Table 5 presents the regression results, including the adjusted $R^2$ of 0.535, indicating a moderate fit for the model. While the dummy variables for sentence types were included to account for systematic differences between neutral and political sentences, their specific coefficients are not central to the interpretation of bias reduction. The key finding remains that the SProp GNN model shows a marked reduction in bias, as evidenced by the negative and significant beta coefficient for the bias factor.

**Table 5.**

*Regression model - testing for reduction in bias*

| | Dependent variable: Bias Reduced SProp GNN Valence Predictions |
|---|---|
| intercept | 53.58*** |
| | (0.96) |
| bias | -0.78*** |
| | (0.20) |
| neutral Sentences | -1.07 |
| | (1.21) |
| political Sentences | -8.62*** |
| | (1.24) |
| Observations | 66 |
| $R^2$ | 0.557 |
| Adjusted $R^2$ | 0.535 |
| Residual Std. Error | 4.04 (df=62) |

*Note.* *p<0.1; **p<0.05; ***p<0.01 (*t*-test)

### Approach 3: Comparing Differences Between Models

The beta coefficient in this analysis explained a significant portion of the variance in valence prediction differences between the two models, with β = 0.78 and a permutation p-value of p = 0.028. This finding strongly supports the conclusion that the SProp model propagates substantially less bias related to political affiliation and gender compared to the transformer model. The positive and significant coefficient indicates that the difference in predictions between the two models is systematically related to the bias factors identified in the transformer model, further highlighting that the SProp GNN effectively reduces the bias originally observed.

Table 6 presents the results of this regression. The intercept (-8.18, $p < 0.001$) represents the baseline difference between the models' predictions, while the bias coefficient ($\beta = 0.78$, $p < 0.01$) accounts for a significant portion of the variance. The inclusion of dummy variables for neutral and political sentences adjusts for systematic differences across stimulus types. While these coefficients (neutral: $\beta = 2.14$, $p < 0.1$; political: $\beta = 7.11$, $p < 0.001$) suggest some variation in the magnitude of prediction differences based on sentence type, the primary finding lies in the bias coefficient itself, which demonstrates the central role of bias reduction in distinguishing the SProp GNN's predictions from those of the transformer model.

**Table 6.**
*Regression model – explaining the difference in predictions*

|  | Dependent variable: |
|---|---|
|  | Difference between Transformer and SProp GNN Valence Predictions |
| intercept | -8.18*** |
|  | (0.94) |
| bias | 0.78*** |
|  | (0.19) |
| neutral Sentences | 2.14* |
|  | (1.19) |
| political Sentences | 7.11*** |
|  | (1.21) |
| Observations | 66 |
| $R^2$ | 0.415 |
| Adjusted R² | 0.386 |
| Residual Std. Error | 4.368 |
|  | (df=62) |

*Note.* *p<0.1; **p<0.05; ***p<0.01 (*t*-test)

The analysis tested three hypotheses to evaluate the bias robustness of the SProp GNN model compared to the transformer model. The null hypothesis about the SProp GNN being robust to bias was retained, but the results were inconclusive. The second hypothesis, which posited that the SProp model's bias is equivalent to the transformer model's, was rejected, showing a significant reduction in bias. The third hypothesis, testing whether prediction differences between the models are related to bias factors, was also rejected, confirming that the SProp model mitigates the biases observed in the transformer model. These results strongly support the conclusion that the SProp GNN substantially reduces bias and is a reliable alternative for unbiased sentiment analysis.

**Explainability**

The SProp model trained on the EmoBank dataset was used to assess the valence of two sentences: "I am happy", and "I am not happy". The former sentence received a valence prediction of 0.68, and the latter, a valence prediction of 0.43 indicating that the model is able to take negation into account and appropriately modify its prediction. Figures 2, and 3 depict an abstracted representation of what happened inside the model during the prediction. The size of the nodes symbolizes the extent to which the model paid attention to them during prediction, and the arrows symbolize edges through which the emotional information was propagated.

*Figure 2. The explanatory graph for sentence "I am happy"*

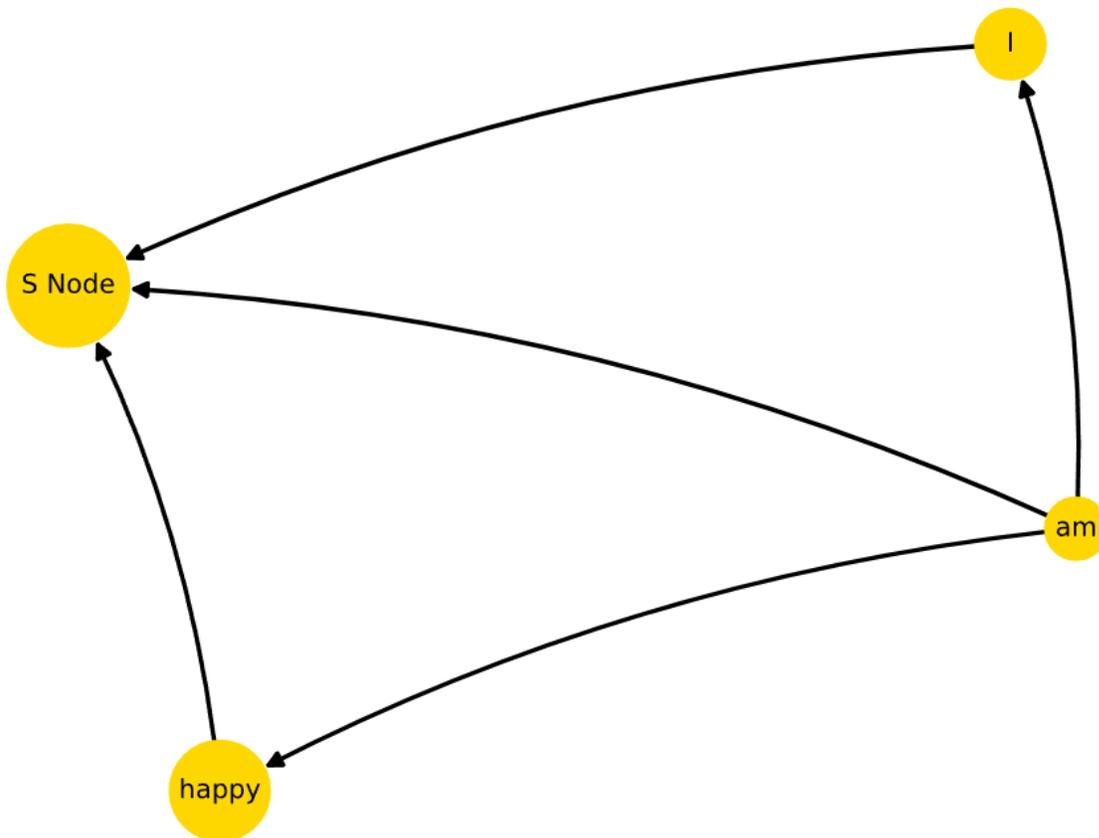

*Note.* The size of the nodes represents the degree to which the model attended to a given node's feature when extracting information from the graph. The S Node refers to the sentence node.

*Figure 3. The explanatory graph for sentence "I am not happy"*

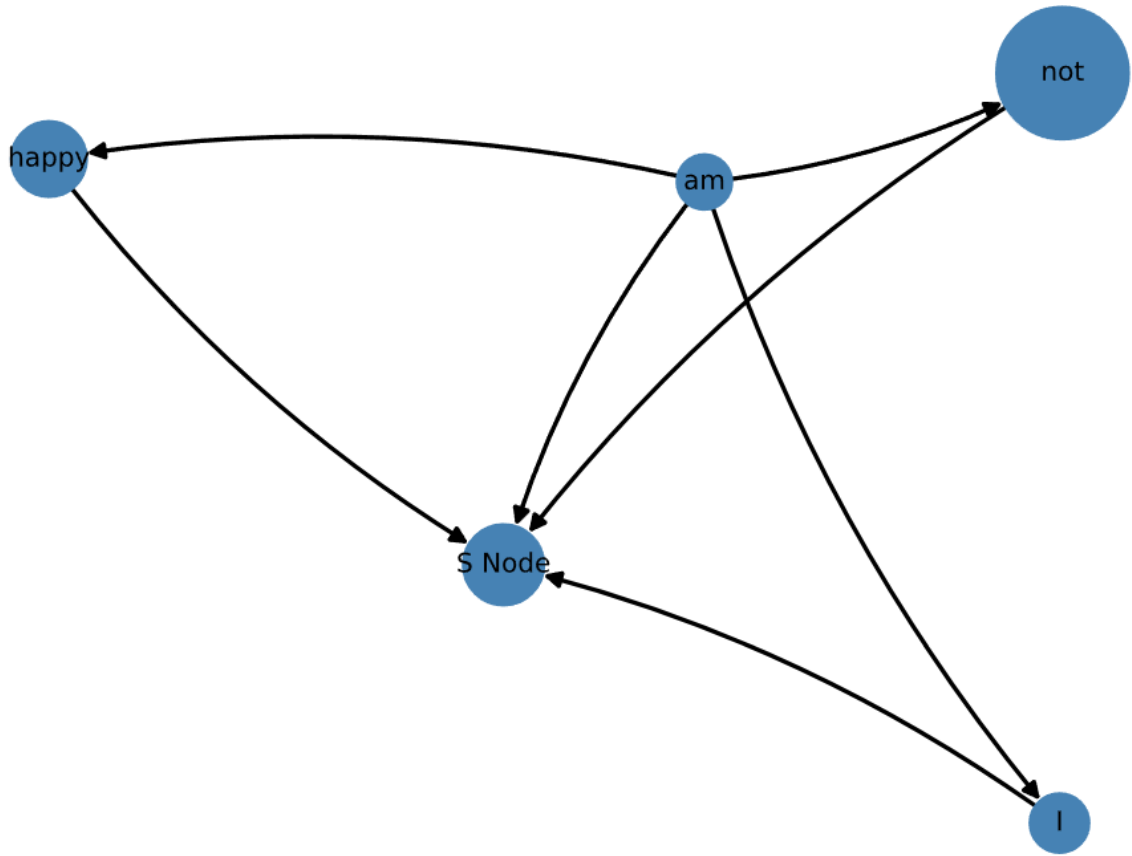

It can be seen that in the case of the first sentence (Figure 2), the model relied on the emotional information from the sentence node, and the node that contained the emotional features of the word "happy". This is expected as the sentence node contains information passed from all of the word nodes in the sentence, while the "happy" node is an adjective, and so often conveys emotional information. In the case of the second sentence, however (Figure 3) the model paid significantly more attention to the "not" node features, indicating that it learned that negation could reverse the emotional load of a sentence.

However, as can be seen in Figure 3, the pathway of emotional propagation from the word "happy" does not include the negation node. This means that the mechanism through which SProp GNN operates is partially based on heuristics, rather than just on the propagation of emotional information through the syntactic graph. To test this, a longer sentence including a negation that should not modulate the emotional information was processed by the model. The sentence "I am happy, and not tall" was given a valence prediction of 0.431, while the same sentence "I am happy, and tall" received a rating of 0.69.

The scaling factors used to propagate the emotional information through the syntactic graph were not visualized, as they do not convey a lot of information. This is due to their high dimensionality. In the current models, the dimension of the scaling factors is equal to the hidden dimension of the SProp Layer – 512. This means that their average in all likelihood obscures a lot of information about their underlying mechanisms.

### Discussion

The present study introduces the Semantic Propagation Graph Neural Network (SProp GNN) as a novel approach to emotion prediction, addressing the critical issue of bias propagation inherent in machine learning based models. The SProp GNN significantly outperformed other lexicon-based alternatives across two different languages—English and Polish—and two distinct emotion prediction tasks, namely discrete and dimensional emotion prediction. It's ability to utilize the syntactic structure of sentences embedded with emotional information on the single word level allowed it to bridge the gap between simple lexicon-based methods and complex black-box models.

The main contribution of this work is the demonstrable reduction of bias in emotion prediction. Such bias can lead to unfair or discriminatory outcomes, both in real world applications such as mental health assessments (Parikh et al., 2019), hiring processes (Kassir et al., 2023), or criminal justice systems (Joseph, 2024), as well as in the academia, where scientific conclusions are required to be fair and objective. The statistical evidence presented shows that the SProp GNN propagates at least significantly less bias than its transformer-based counterpart. This evidence, coupled with the sole fact that the SProp GNN simply does not have access to any bias information it could overfit, since it only processes the syntactic structure of the sentence coupled with external emotional ratings and not words directly, warrants a claim that it is robust to training data bias. By effectively mitigating bias, the SProp GNN not only enhances the fairness and ethical standing of emotion prediction tools but also increases their reliability across diverse populations. This is particularly crucial in a global context where texts may reflect a wide array of cultural, social, and individual differences. The ability of the SProp GNN to provide more objective emotional assessments can contribute to more equitable decision-making processes in applications that rely on sentiment analysis.

While the SProp GNN performs slightly worse than pretrained transformers, it constitutes a viable alternative in applications where objective emotional assessment is of key importance. The development of the SProp GNN highlights the inherent trade-offs between model interpretability, performance, and bias mitigation. Transformer-based models often achieve higher accuracy due to their ability to capture complex semantic nuances; however, they also tend to act as black boxes, making it difficult to understand or control the sources of their predictions, including biases. In contrast, the SProp GNN offers a more interpretable architecture allowing for greater transparency in how predictions are made. Although there is a slight decrease in performance compared to transformers—a mean accuracy difference of 5.70 percentage points on the GoEmotions dataset, for instance—this trade-off may be acceptable or even preferable in contexts where interpretability and bias reduction are prioritized over marginal gains in accuracy. This balance underscores the importance of aligning model selection with the specific requirements and ethical considerations of the intended application.

The bias robustness of the SProp GNN makes it particularly suitable for applications where fairness and objectivity are paramount. For instance, in the analysis of social media data for public health monitoring (Babu & Kanaga, 2022), using a model that minimizes bias ensures that interventions are based on accurate representations of population sentiments without skewing toward or against specific groups. In mental health contexts, such as suicide risk prediction from text messages (Glenn et al., 2020), unbiased sentiment analysis can lead to more accurate assessments and timely interventions. Increasingly popular tools that analyze student feedback (Dalipi et al., 2021) can also benefit from unbiased emotion assessments to foster an inclusive learning environment. In all these cases and more, the SProp GNN's ability to deliver high-performance emotion predictions while mitigating bias is of significant practical value.

**Limitations and Future Research**

The SProp GNN seems susceptible to simplistic heuristics, as shown in the explainability section, where the model did not fully capture the nuanced role of negations in complex sentences. This potential discrepancy between the good performance results of the model and its inability to pick up on syntactic cues that are easily understandable to a human can be explained by the low frequency of such sentence structures in the tested datasets and their overall rarity. The datasets used are diverse in language and task types, but they may not encompass the full spectrum of linguistic structures and expressions found in real-world texts. This could limit the generalizability of the SProp GNN to other languages or dialects not represented in the training data.

Future studies can further improve the architecture of the SProp GNN, shrinking the gap between its performance and that of other potentially biased models. One potential avenue for further exploration could be a modification of the syntactic graph creation algorithm. While the syntactic pathways extracted using the spaCy package (Ines Montani et al., 2023) provide useful information about the structure of sentences, a more tailored model that would map the pathway of emotional information propagation directly could achieve even better results, potentially reducing the model's reliance on heuristics. Expanding the training datasets to include more syntactically diverse sentences could help the model learn to handle complex linguistic structures more effectively.

Furthermore, the model's reliance on lexicons, or norm-extrapolation models could introduce bias present on the word level. Despite the lack of context when annotating emotions at word level, there is still a small possibility that the annotators for the lexicons impacted some sorts of biases on the emotion dictionary. This type of bias, however, can usually be easily explored using the lexicon in question, and its mitigation is as simple as equalizing the emotional load of words that convey it. In circumstances where a specific type of bias could directly impact the conclusions of a study, checking the lexicon for its presence before the use of the SProp GNN is advisable.

## Expanding the Concept of Semantic Blinding

The proposed approach of selectively withholding specific semantic information from the model, termed semantic blinding, is a technique that deliberately limits the model's access to particular semantic details. By preventing the model from associating emotional predictions with specific words or concepts that could introduce unwanted biases, semantic blinding ensures that the model's emotional assessments are free from training data biases related to specific groups or subjects. This technique presents exciting opportunities for future research. It could be extended to other natural language processing tasks where bias could be a concern, such as text-classification. Exploring how semantic blinding can be integrated with transformer-based architectures might also yield models that combine the high performance of transformers with the bias mitigation benefits of the SProp GNN. Additionally, further investigation into the types of semantic information that can be withheld without significantly impacting performance could lead to the development of more robust and fair NLP models across various domains.

## Practical Considerations for Deployment

From a practical standpoint, deploying the SProp GNN in real-world applications offers significant advantages in terms of computational efficiency and scalability due to its substantially smaller model size when compared to its transformer-based counterparts. Specifically, the SProp GNN trained on the EmoBank dataset consists of approximately 1.5 million parameters, while the transformer model trained for the same task comprises about 125 million parameters. This significant reduction in model size—over 20 times smaller—translates to lower computational overhead and faster processing times, making the SProp GNN more suitable for deployment on devices with limited resources or for applications requiring real-time analysis. For such applications, the word level prediction stage of the model could be done prior to inference time by generating a very large emotional dictionary a priori. From an academic standpoint, this translates to accessibility for researchers without access to high-performance computing resources. The reduced memory and processing requirements mean that the SProp GNN can be trained and deployed on standard hardware, broadening the scope of researchers who can experiment with and apply this model. This adaptability and efficiency make the SProp GNN a practical and accessible alternative to transformer-based models in sentiment analysis tasks.

In order to allow other researchers to replicate the analyses presented in the current paper and use the SProp GNN architecture for their research, the code, along with detailed comments for this paper has been made available at a GitHub repository (https://github.com/hplisiecki/Semantic-Propagation-GNN). Additionally, the Technical Appendix should serve as additional guide for those willing to apply and further develop the methods here presented.

## Conclusion

The SProp GNN represents a significant step forward in developing sentiment analysis models that prioritize fairness and interpretability without substantial sacrifices in performance. The evidence demonstrates that the SProp GNN not only approaches the accuracy of transformer-based models but also at least significantly reduces the propagation of biases. This coupled with the fact that the model does not possess the ability to overfit specific words points towards near total bias eradication. By addressing the critical issue of bias propagation, the SProp GNN offers a viable and ethically sound alternative for a wide range of applications. While there is room for improvement, particularly in handling complex syntactic structures and expanding language coverage, the SProp GNN lays the groundwork for future advancements in unbiased and interpretable sentiment analysis. Future work focused on enhancing the model's architecture, expanding its applicability, and refining the semantic blinding technique holds the promise of further bridging the gap between high performance and bias mitigation in natural language processing

**Funding**

This research is funded by a grant from the National Science Centre (NCN) 'Research Laboratory for Digital Social Sciences' (SONATA BIS-10, No. UMO-020/38/E/HS6/00302).

## Technical Appendix

This technical appendix provides detailed information on the methodologies, models, datasets, and experimental setups used in the paper. It is intended to offer in-depth insights that supplement the main text, as well as to serve as a guide for training similar models in the future.

## Detailed Emotion Prediction Pipeline

The GNN model proposed in the manuscript consists of three stages:

1. Word level emotion prediction
2. Syntactic graph creation
3. The Semantic Propagation GNN (SProp GNN)

Below each of these stages are described in detail

### Word Level Emotion Prediction

The model relies on knowing the emotions of every, or most words in a text, to then propagate this information through the syntactic graph and predict emotion on the text level. The paper, in order to predict the emotions of words, draws on the literature in norm extrapolation (Plisiecki & Sobieszek, 2023) which recommends the use of transformer models for word level emotion prediction. The use of these models might not be necessary given a large enough lexicon of words and their respective emotional values, and a corpus with restrained vocabulary. However, to ensure proper word coverage already trained transformer norm extrapolation models are used, or, in the case of discrete emotion prediction, new ones are trained

Transformer based norm extrapolation models are trained by adding a regression or a classification head to a transformer encoder and training it on an existing norm lexicon. Previous research has also added an additional hidden layer between the encoder, and the regression head, with dropout. To create one for the task of discrete emotion prediction the NRC Emotion Intensity Lexicon was used (Mohammad, 2017) which provides ratings on a scale of 0 to 1 for 5891 unique words for eight emotions (anger, anticipation, disgust, fear, joy, sadness, surprise, trust). Not all of the words were rated with regards to each emotion, which required training eight separate prediction models for each of the emotions. The "nghuyong/ernie-2.0-en" model was used for each of them, as this was the model that was also used for valence and arousal prediction in a previous paper and thus can be trusted to model emotional information well (Plisiecki & Sobieszek, 2023). The lexicon was split into seven emotion specific lexicons, and each of these subcorpora was then further divided into train, evaluation, and test sets in the ratio of 8:1:1. Each model was trained for 100 epochs, with a batch size of 500, learning rate of 5e-5, AdamW optimizer with a weight decay of 0.3 and a linear learning rate schedule with warmup steps amounting to 600, a hidden 768 dimensional hidden layer and a dropout of 0.1. Early stopping based on the correlation of predicted scores with the ground truth on the validation set was implemented to prevent overfitting. The test set correlations for each of the emotions are presented in Table 1.

Table 1.

*Discrete Emotion Norm Extrapolation Model Performance (Pearson's Correlations)*

| Emotion | Anger | Anticipation | Disgust | Fear | Joy | Sadness | Surprise | Trust |
|---------|-------|-------------|---------|------|-----|---------|----------|-------|
| Correlation | 0.77*** | 0.68*** | 0.73*** | 0.74*** | 0.71*** | 0.71*** | 0.81*** | 0.72*** |

* p < 0.6, ** p < 0.1, *** p < 0.001

The performance metrics of already existing valence and arousal models for Polish and English taken from Plisiecki and Sobieszek (2023) are reported in Table 2.

Table 2.

*Continuous Norm Extrapolation Model Performance for Polish and English (Pearson's Correlatrions)*

| Language | English | Polish |
|----------|---------|--------|
| Valence | 0.95*** | 0.93*** |
| Arousal | 0.76*** | 0.86*** |

* p < 0.6, ** p < 0.1, *** p < 0.001

In the pipeline these models were used to assess the emotional value of all words that weren't stop words, punctuations, or negations, as assessed by the *spaCy* package (Ines Montani et al., 2023). To improve the compute time of the emotion prediction pipeline, using similar models to create a very big lexicon prior to inference can be an option.

**Syntactic Graph Creation**

The current pipeline uses the spaCy package (Ines Montani et al., 2023) to split the text into sentences, and words, followed by an analysis of syntactic dependencies. Each word is connected to the other words it relates to syntactically. For example, in the sentence "I do not feel well," spaCy identifies "feel" as the main verb, with "I" as its subject and "well" as its modifier. Additionally, the negation "not" is linked to "feel," indicating a negative sentiment in the phrase. This information can be represented in a graph form where nodes are words, and edges are syntactic dependencies. Each word is furthermore assigned to a specific part-of-speech category (e.g., PRON (pronoun) for "I" and VERB (verb) for "feel") and each dependency labeled accordingly (e.g., nsubj (nominal subject) for "I" as the subject of "feel" and neg (negation) for "not" modifying "feel").

All of punctuation marks are removed from the text, prior to the construction of the graph, apart for the ellipsis, exclamation, and question marks ('…', '! ', '? ') which were retained as they play a big role in the modulation of text meaning. While spaCy recognizes only around 20 part of speech tags, its taxonomy for dependency types is much larger. For this reason, they have been recategorized to a more manageable taxonomy of 15 separate categories with entries like "Descriptive Modifiers of Verbs", or "Negations". The full mapping is available on the paper's github repository (https://github.com/hplisiecki/Semantic-Propagation-GNN).

The resulting structure is a graph where the nodes (words) are assigned feature vectors with the emotion ratings predicted at the word level emotion prediction stage, along with a number signifying their position in the sentence (word index divided by the number of words in the sentence). Words are also assigned parts of speech indexes signifying their parts of speech categorization. Finally, each of the edges (connections) within the graph gets assigned their dependency indexes, relating to the dependency type taxonomy.

In order to model not only single sentences but also multiple sentences texts, all words from a sentence are additionally to an additional sentence node. When a text has more than one sentence, the sentence nodes relating to each sentence get connected to each other sequentially in the order they appear in text. These sentence nodes are "empty" in the sense that they are not assigned any emotional information. Instead, their emotion node features are initialized at zero, allowing the graph to propagate the emotion from words into them at inference. Their node features also contain their sentence number indicator (sentence index divided by the number of sentences in the text). Finally, they are also assigned a unique parts of speech category (the same for every sentence), with their edges having a unique dependency category (the same for every sentence).

**Semantic Propagation Graph Neural Network**

The SPROP GNN rests on the idea of allowing the model to propagate semantic information, in the form of word sentiment scores throughout the syntactic graph as part of the inference. It is able to do it thanks to the custom SPROPConv layer which considers information about the parts of speech each of the two words (nodes) connected in the graph belong to, the emotional information of the receiving node as well as the type of syntactic dependency (edge) between them.

Below is a general overview of the steps that the model performs, followed by a more formal explanation of how the SPROPConv layer works, and a short description of the rest of the model's architecture. Because this paper introduces the SPROPConv layer, much attention is paid to its description. Afterwards, the training setup is described.

***General Steps Performed by the SPROP GNN***

1. **Process Syntactic Graph with SPropConv Layer:**

    The model processes the syntactic graph of the text using the custom SPropConv layer, enriching each word's representation with information from related words based on their grammatical structure and roles.

2. **Concatenate with POS Embeddings:**

    Each word's updated features are combined with its part-of-speech (POS) embedding, adding grammatical context to each word's representation within the graph.

3. **Apply Attention Pooling:**

    The concatenated embeddings are passed to an attention pooling layer, which identifies and weighs the most relevant words in the graph for predicting the text's emotional tone. These weighted embeddings are then aggregated using a global addition pool to create a cohesive text representation.

4. **Pass Through Fully Connected Layers:**

The pooled text representation is further processed through fully connected layers. These layers refine and adjust the representation to reach the dimensionality needed for the final prediction.

**5. Generate Final Prediction:**

For continuous emotional metrics, the output layer uses sigmoid activation to predict values between 0 and 1 for each metric. For discrete emotion categories, a softmax activation generates probabilities across categories, identifying the most likely emotion.

*Mathematical Formulation of the SPropConv Layer*

The SProp layer operates through a series of steps that involve transforming node features, computing messages between nodes, and updating node representations.

**1. Node Feature Transformation**

Each word in the sentence is initially represented by a feature vector which consists of the emotional score of each word, alongside its index sentence divided by sentence length. These features are transformed to a hidden representation using a linear transformation:

$$h_i = W_x x_i + b_x$$

- $h_i$: Hidden representation of node iii.

- $W_x, b_x$: Learnable parameters (weights and biases).

**2. Message Passing**

For each edge from node $j$ to node $i$ (representing a syntactic dependency), the model computes a message that incorporates:

- The hidden representation of the source node $h_j$

- The embeddings of the POS tags for both nodes: $t_i$ for node $\boldsymbol{i}$ and $t_j$ for node $\boldsymbol{j}$.

- The embedding of the edge type (syntactic dependency) $s_{ij}$

These components are concatenated and passed through a linear transformation followed by a hyperbolic tangent activation (tanh) to compute a scaling factor $s_{ij}$

$$s_{ij} = \tanh\left(W_s[h_j; t_i; t_j; e_{ij}] + b_s\right)$$

- $[\cdot;\cdot]$ : Concatenation operation.

- $W_s, b_s$: Learnable parameters.

**3. Message Computation**

The message from node $j$ to node $i$ is calculated by scaling the hidden representation of node $j$ with the scaling factor $s_{ij}$:

$$m_{ij} = \boldsymbol{s_{ij}} \cdot h_j$$

This step allows the model to modulate the influence of node $j$ on node $i$ based on their syntactic and semantic relationship.

### 4. Aggregation

For each node $i$, the incoming messages from all its neighboring nodes are aggregated using summation:

$$a_i = \sum_{j \in \mathcal{N}(i)} m_{ij}$$

- $\mathcal{N}(i)$: Set of neighboring nodes of node iii.

### 5. Update

The node's hidden representation is updated by combining its original hidden state with the aggregated messages, followed by a rectified linear unit (ReLU) activation:

$$h_i' = \textbf{ReLU}(h_i + a_i)$$

- $h_i'$: Updated hidden representation of node iii.

### *The Remaining Architecture*

After the syntactic graph has been processed using the SPropConv layer, the model concatenates the graph's matrix representation with the parts-of-speech embeddings for each word in the syntactic graph. This concatenated embedding is then passed to an attention pooling layer, which identifies the words in the graph that contain the most relevant information for predicting the text's emotional tone, assigns them weights, and aggregates these embeddings using a global addition pool.

This representation is then passed through fully connected layers that gradually bring them to the dimensionality required by the prediction. In the case of continuous emotional metrics, this means one output dimension per predicted metric, with a sigmoid activation applied to scale the output between 0 and 1. For discrete emotion prediction, the final layer instead uses a softmax activation, outputting probabilities across predefined emotion categories.

### *Specific Architectures and Training Setup*

The three SProp GNN models trained on the three datasets GoEmotions, EmoBank, and the Polish Political Dataset share a similar architecture. Each model contains a single SProp layer with 512 hidden dimensions, alongside embedding layers for both parts of speech (node types) and dependency relationships (edge types), with dimensions matching those of the SProp layer. This is followed by a global attention mechanism, which applies a gated attention layer configured with two linear transformations (1024 to 256, and 256 to 1) and a ReLU activation in between. The attention weights are computed by applying softmax across nodes within each graph, and graph-level features are subsequently aggregated using a global addition pool.

The differences between the models lie in the final sequence of linear layers. In the case of discrete predictions, these layers have the form of three linear transformations (1024 to 1024, 1024 to 512, and 512, to the number of discrete emotions), separated by dropout and relu activations. Alternatively, in the case of the two continuous metric prediction models there are only two linear layers (1024 to 100, and 100 to 1), also separated by a dropout and a relu activation. These differences stem from free experimentation with different amount of final linear layers. A systematic exploration of alternative architectural setups is beyond the scope of this study.

The hyperparameters for the three SProp GNN models were chosen using a Bayesian hyperparameter sweep on the wandb platform (*Wandb/Wandb*, 2017/2024). The hyperparameter options for the three models were the same: dropout - 0, 0.2, 0.4, 0.6; learning rate - 5e-3, 5e-4, 5e-5, and weight decay - 5e-3, 5e-4, 5e-5. All models were trained using the AdamW optimizer with the epsilon equal to 1e-6 and betas equal to 0.9, and 0.999. The discrete model used the cross-entropy loss, while the continuous metric prediction models used the mean squared error loss. The final models were trained using the best performing parameters from the sweeps.

## Comparative Experiments

This section will outline the data wrangling performed on the datasets that were used to compare the SProp GNN with other methods, along with the explanation of how each of the alternative methods were implemented.

### Data Wrangling

Each of the datasets was processed for the task of using them to compare alternative approaches to emotion prediction. Considerable attention was paid to the description of the Polish political dataset as it is a far less known dataset when compared to the other two.

#### *The Goemotions Dataset*

The goemotions dataset was developed by a team at Google (Demszky et al., 2020). It consists of 57565 unique texts and 210622 annotations. Each comment received annotations from three English-speaking raters from India, with additional raters assigned when agreement was low. The most voted for emotion per each text was computed and those texts for which two emotions were assigned the same number of votes were dropped. This resulted in a dataset of 47136 unique texts. From these texts, those that were assigned one of the following emotions: anger, disgust, fear, joy, surprise; were retained leaving 4819 unique texts. The choice of emotions was dictated by the availability of emotion norms in the NRC Emotion Intensity Lexicon (Mohammad, 2017). This dataset was then split into the training, evaluation, and test sets in the proportion of 8:1:1.

#### *The Emobank Dataset*

The EmoBank dataset, created by Buechel and Hahn (2022), consists of 10,062 English sentences from sources like news, blogs, fiction, and letters, annotated along three emotional dimensions: Valence,

Arousal, and Dominance (VAD). Each sentence was rated by multiple annotators from the crowdsourcing platform CrowdFlower for both *writer* and *reader* perspectives, giving insights into both expressed and perceived emotions. Each sentence was annotated by 5 annotators. In accordance with the recommendations of the researchers, the dataset with the weighted average of the reader and writer perspective labels provided at their online repository was used for training (*JULIELab/EmoBank*, 2017/2024). The ratings for valence and arousal were normalized to a 0 to 1 range by subtracting the lowest score and dividing by the number of Likert scoring options prior to splitting into the training, evaluation, and test sets in the proportion of 8:1:1.

### *The Polish Political Dataset*

The Polish Political dataset (Plisiecki et al., 2024) was created by sampling text data from social media profiles of Polish journalists, politicians, and non-governmental organizations (NGOs) across YouTube, Twitter, and Facebook. Posts from 2019 onward were collected for 69 profiles. A total of 1,246,337 text snippets were gathered, with breakdowns of 789,490 tweets, 42,252 YouTube comments, and 414,595 Facebook posts. To handle the varying text lengths, Facebook posts were split into sentences, and only texts under 280 characters were retained. Social media artifacts, such as dates and extraneous links, were removed, and non-Polish texts were filtered using language detection software. To prevent overfitting, online links and usernames were standardized as "*link*" and "*user*."

To create a dataset with richer emotional content, neutral texts were filtered out, leaving only those with higher levels of emotional valence, arousal, and dominance. This selection process used a lexicon-based approach, where each text was assessed for emotional intensity across these dimensions, resulting in 8,000 emotionally charged texts. An additional 2,000 neutral texts were included to balance the dataset, preserving original platform proportions. The final 10,000-text dataset, comprising 496 YouTube comments, 6,105 tweets, and 3,399 Facebook texts, was then annotated by 20 psychology students well-versed in Polish political discourse. Each text was rated by five randomly assigned annotators on six emotions (happiness, sadness, disgust, fear, anger, and pride) and two emotional dimensions (valence and arousal), using a 5-point Likert scale. Before formal annotation, annotators received an introduction to valence and arousal, and comprehensive guidelines were provided to ensure consistency. For clarity, the annotators received the following English instruction for evaluating valence and arousal:

"Go back to the text you just read. Now think about the sign of emotion (positive / negative) and the arousal you read in a given text (no arousal / extreme arousal). Rate the text on these emotional dimensions."

This instruction was designed to provide a standardized understanding of emotional dimensions, ensuring alignment in annotators' assessments across the dataset.

For the purposes of the current experiment, all of the emoticons and symbols were filtered out and the dataset was split into the training, evaluation, and test sets in 8:1:1 proportion.

### Comparative Approaches

This section outlines the details of how each of the alternative approaches was set up and trained for performance comparison.

***The Lexicon Approach***

For the lexicon analysis of the EmoBank, and the Polish Political dataset I utilize the norm extrapolation transformer-based models for Polish and English described in the "Word Level Emotion Prediction" section above. Each test set text was first split into words using the *spacy* package (Ines Montani et al., 2023). Each word that wasn't a stop word was then fed into the norm extrapolation model, and the emotional prediction was averaged to get the text level emotion score.

***The Vader Approach***

VADER (Valence Aware Dictionary and Sentiment Reasoner) is a rule-based model designed for sentiment analysis, particularly effective in capturing sentiment from social media and informal text. I have used the 3.3.2 version of the Vader package to get the valence/positivity scores for the Emobank test set.

***The EmoAtlas Approach***

The EmoAtlas utilizes an extensive lexicon-based network to profile emotions by mapping syntactic and semantic relationships in text, effectively capturing nuanced emotional cues without extensive model training. The EmoAtlas performance results for the goemotion dataset were taken directly from the original paper (Semeraro et al., 2023).

***The Transformer Approach***

For the Goemotion and EmoBank datasets the *roberta-base* transformer model developed by Facebook was finetuned on the two English datasets (Liu et al., 2019). A fully connected layer, with the dimensions equal to 768 was added on the top of the base model with dropout and a layer norm, with either a regression head for the sake of predicting valence and arousal, or a classification head for predicting discrete emotions. A Bayesian hyperparameter sweep was performed using the wandb platform (*Wandb/Wandb*, 2017/2024) for both models with 20 runs and the following hyperparameter options: dropout – 0.0, 0.2, 0.4, 0.6; learning rate – 5e-4, 5e-5, 5e-6; weight decay – 0.0, 0.2, 0.4, 0.6; and warmup steps – 300, 600, 900. Both models use the AdamW optimizer for training with the epsilon equal to 1e-6 and betas equal to 0.9, and 0.999, alongside the linear learning rate scheduler with warmup. In the case of discrete prediction, cross-entropy loss was used, while in the case of continuous emotion metric prediction mean squared error loss was chosen. Finally, the final models have been trained using the best performing hyperparameters from the sweep. The performance of each of the models is reported in the results section of the main manuscript. The training code for these models can be found in the following Google Collab:

https://colab.research.google.com/drive/1pA3oBbHg0pza1yF5kyuddK36-RGKtHxo?usp=sharing